\documentclass{article}

\usepackage{iclr2026_conference,times}
\usepackage{url}
\iclrfinalcopy 

\usepackage{microtype}
\usepackage{graphicx}
\usepackage{subfigure}
\usepackage{booktabs}
\usepackage{colortbl}
\usepackage{xcolor}
\usepackage{tcolorbox}
\tcbuselibrary{most}
\usepackage{hyperref}
\usepackage{amsmath,amssymb,mathtools,amsthm}
\usepackage{enumitem}
\usepackage{tabularx}
\usepackage[capitalize,noabbrev]{cleveref}
\usepackage{adjustbox}
\usepackage{algorithm}
\usepackage{algpseudocode}
\usepackage{hyphenat}

\theoremstyle{plain}

\theoremstyle{definition}

\theoremstyle{remark}

\definecolor{lightpurple}{RGB}{221,160,221}
\definecolor{grey}{RGB}{128,128,128}
\definecolor{lightblue}{RGB}{210,225,250}

\newcommand{\model}{\texttt{StepORLM}}

\usepackage{listings}
\usepackage{xcolor}

\definecolor{codebg}{rgb}{0.95,0.95,0.95}
\title{StepORLM: A \underline{S}elf-Evolving Framework With Genera\underline{t}iv\underline{e} \underline{P}rocess Supervision For \underline{O}perations \underline{R}esearch \underline{L}anguage \underline{M}odels}

\author{
    \parbox[t]{0.45\textwidth}{
        \normalfont
        \textbf{Chenyu Zhou} \\
        Shanghai Jiao Tong University \\
        Shanghai, China \\
        \texttt{chenyuzhou@sjtu.edu.cn}
    }
    \hspace{0.05\textwidth} 
    \And
        \parbox[t]{0.45\textwidth}{
        \normalfont
        \textbf{Tianyi Xu} \\
        Shanghai Jiao Tong University \\
        Shanghai, China \\
        \texttt{crimsonflag@sjtu.edu.cn}
    }\\
    \And
    \vspace{1.5\baselineskip}
        \parbox[t]{0.45\textwidth}
        {
        \normalfont
        \textbf{Jianghao Lin\thanks{Corresponding author.}} \\
        Shanghai Jiao Tong University \\
        Shanghai, China \\
        \texttt{linjianghao@sjtu.edu.cn}
        }
    \hspace{0.05\textwidth}
    \And
    \parbox[t]{0.45\textwidth}{ 
    \normalfont
    \textbf{Dongdong Ge} \\
    Shanghai Jiao Tong University \\
    Shanghai, China \\
    \texttt{ddge@sjtu.edu.cn}
    }
}

\begin{document}
    \maketitle

\begin{abstract}

Large Language Models (LLMs) have shown promising capabilities for solving Operations Research (OR) problems. 
While reinforcement learning serves as a powerful paradigm for LLM training on OR problems, existing works generally face two key limitations. First, outcome reward suffers from the \textit{credit assignment problem}, where correct final answers can reinforce flawed reasoning.
Second, conventional discriminative process supervision is \textit{myopic}, failing to evaluate the interdependent steps of OR modeling holistically. 
To this end, we introduce \textbf{\model{}}, a novel self-evolving framework with generative process supervision. 
At its core, \model{} features a co-evolutionary loop where a policy model and a generative process reward model (GenPRM) iteratively improve on each other. 
This loop is driven by a dual-feedback mechanism: definitive, outcome-based verification from an external solver, and nuanced, holistic process evaluation from the GenPRM. 
The combined signal is used to align the policy via Weighted Direct Preference Optimization (W-DPO) and simultaneously refine the GenPRM. 
Our resulting 8B-parameter \model{} establishes a new state-of-the-art across six benchmarks, significantly outperforming vastly larger generalist models, agentic methods, and specialized baselines. 
Moreover, the co-evolved GenPRM is able to act as a powerful and universally applicable process verifier, substantially boosting the inference scaling performance of both our own model and other existing LLMs. 
\end{abstract}

\section{Introduction}
Large language models (LLMs) have shown remarkable capabilities for complex reasoning, enabling new frontiers in specialized domains like Operations Research (OR)~\citep{xiao2025survey}. Research on applying LLMs to OR has largely advanced along two fronts: 1) developing agentic systems~\citep{optimus0.3, xiao2024chainofexperts} and 2) improving LLMs through specialized training~\citep{tang2024orlm}. Within the latter, reinforcement learning (RL) has become a dominant paradigm through outcome reward~\citep{chen2025solverinformedrlgroundinglarge} or process supervision~\citep{Astorga2024,chen2025stepwiseguidedpolicyoptimization}.

\begin{figure}[t]
    \centering
    \includegraphics[width=1\linewidth]{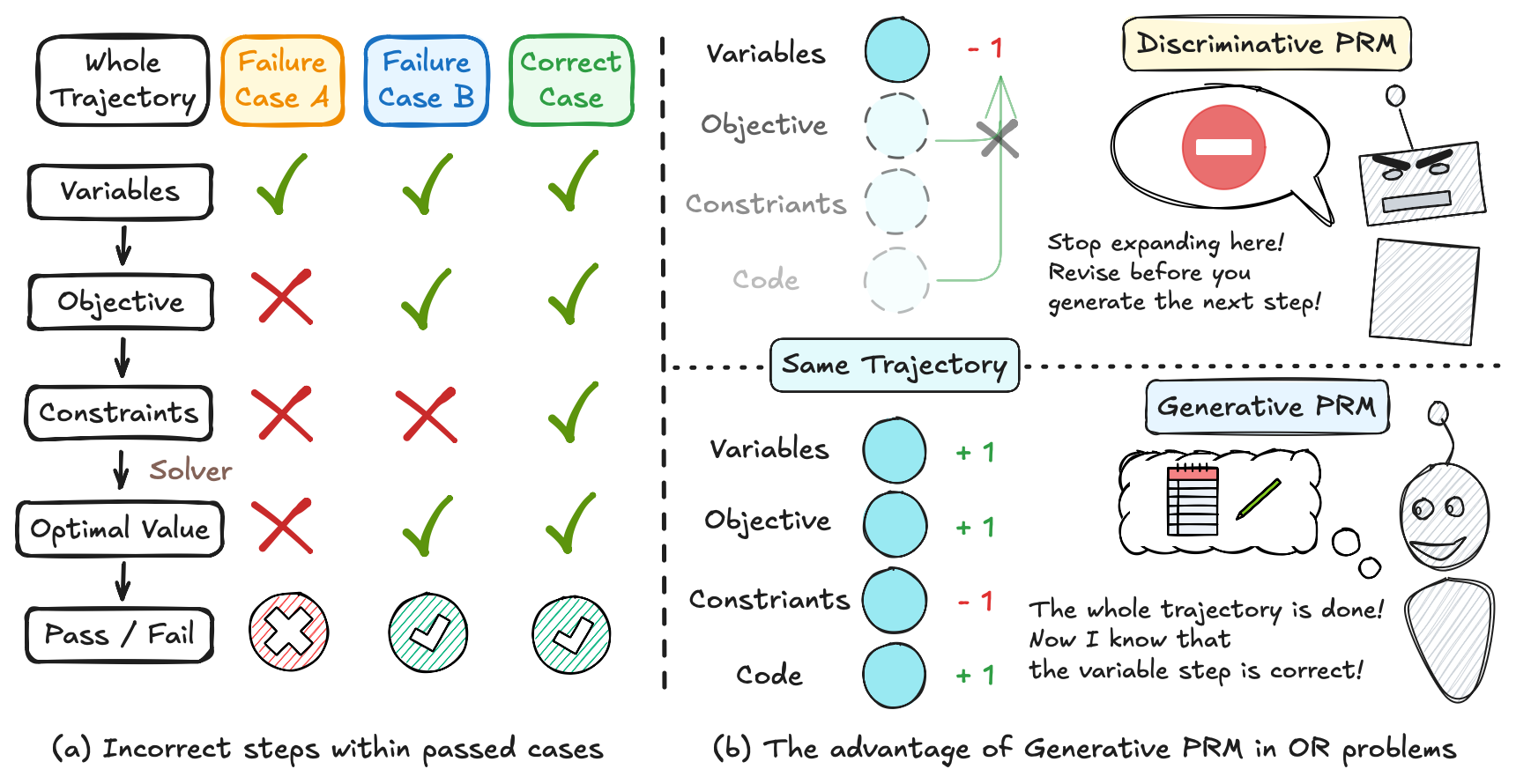}
    \vspace{-15pt}
    \caption{The illustration of (a) the credit assignment problem of outcome reward and (b) the myopic issue of discriminative process supervision, which motivates the application of generative process supervision for operations research language models.}
    \vspace{-15pt}
    \label{fig:opening-figure}
\end{figure}

However, these established RL paradigms exhibit critical flaws when applied to operations research, which requires long-horizon, interdependent reasoning. 
Outcome-based supervision, which rewards a model based solely on the final solution's correctness, suffers from the \textbf{credit assignment problem}. As illustrated in Figure~\ref{fig:opening-figure}(a), an optimal outcome can emerge from a flawed reasoning process, inadvertently reinforcing incorrect intermediate steps. 
While process supervision aims to resolve this by rewarding intermediate steps, conventional discriminative process reward models (PRMs) are often \textbf{myopic} and overlook the critical interdependencies among sequential steps. 
They evaluate each step in isolation and thereby fail to capture the holistic context of an OR solution. 
As shown in Figure~\ref{fig:opening-figure}(b), the correctness of an early step, like a variable definition, may only become apparent once subsequent constraints are defined, causing discriminative PRMs to provide unreliable or inconsistent rewards.

Therefore, we argue that the contextual and interdependent nature of OR modeling demands a paradigm shift from myopic, step-wise evaluation to \textbf{holistic, trajectory-level process supervision}. 
The critic should assess the entire reasoning path retrospectively and understand the intricate dependencies between steps before assigning credit. 
As illustrated in Figure~\ref{fig:opening-figure}(b), this motivates the application of \emph{generative} PRMs that can reason about the completed trajectory and generate a holistic critique, rather than a discriminative one that provides disconnected, localized scores.

To this end, we propose a novel \underline{S}elf-Evolving Framework with Genera\underline{t}iv\underline{e} \underline{P}rocess Supervision for \underline{O}perations \underline{R}esearch \underline{L}anguage \underline{M}odels (\textbf{\model}).
At the heart of our framework is \textbf{a co-evolutionary loop} where a policy model and a novel generative process reward model (GenPRM) iteratively improve each other. 
At each iteration, the policy generates reasoning trajectories that are evaluated from two complementary perspectives: definitive outcome verification from an external solver and nuanced, holistic process feedback from GenPRM. 
This dual-feedback signal is distilled into 1) preference pairs to align the policy via a Weighted Direct Preference Optimization (W-DPO) objective and 2) filtered trajectory rewarding data to further fine-tune GenPRM. 
In this way, both the policy model and GenPRM would be iteratively improved through the self-evolving loop, ultimately leading to process-sound and outcome-correct reasoning capabilities to solve complex OR problems.

Our contributions are summarized as follows:
\begin{itemize}[leftmargin=15pt]
    \item To the best of our knowledge, we are the first to introduce \textbf{generative process supervision} to holistically evaluate the entire reasoning trajectories on OR problems, mitigating the \textit{credit assignment problem} and \textit{myopia issue} of existing RL-based works.

    \item We propose \textbf{\model}, a novel self-evolving framework centered on the co-evolution of a policy model and a generative process reward model (GenPRM). 
    It integrates a scalable data synthesis pipeline with a dual-feedback loop that grounds the models in both process-level logical consistency and final-outcome correctness.

    \item Extensive experiments on seven benchmarks show that \model{} achieves the \textbf{SOTA} performance against both vastly larger generalist/agentic models (e.g., \texttt{GPT-4o}) and specialized baselines. 
    We also demonstrate that the co-evolved GenPRM acts as a \textbf{universal inference-time verifier}, substantially boosting the performance of other OR language models. 

    \item We release our code, the StepORLM model weights, and the GenPRM verifier weights to the community, providing a strong foundation for future research in LLM-based OR problem solving.
\end{itemize}

\section{Related Work}

\paragraph{Large Language Models for Optimization Modeling.}

Early works explore using LLMs to parse natural language descriptions into mathematical models. For example, the NL4Opt competition \citep{Ramamonjison2023} demonstrates that ChatGPT and other LLMs can identify problem entities and generate LP formulations from word problems. Pipeline frameworks \citep{tang2024orlm} combine LLM components with traditional solvers: AutoFormulation \citep{tang2024orlm, Jiang2025a, lu2025optmathscalablebidirectionaldata, zhou2025autoformulatingdynamicprogrammingproblems} uses LLMs to tag variables and constraints in text descriptions, and systems like OptiMUS\citep{AhmadiTeshnizi2024} pipeline textual inputs through LLM-to-code translators before invoking MILP solvers. These methods improve accessibility, but often produce hallucinated constraints or missing logic, as the LLMs have no built-in notion of optimization semantics. Agentic methods~\cite{zhang2024agentic,yang2025survey} are also proposed, which assign specialized LLM agents under a conductor that orchestrates and reflects on partial solutions \citep{xiao2024chainofexperts, deng2024cafa, optimus0.3}. While these approaches outperform simpler pipelines on benchmarks, they still rely on repeated LLM exchanges without explicit verification of each reasoning step. In practice, even advanced models exhibit unsound reasoning: for instance, \cite{Jiang2025a} note that alignment and “self-correction” mechanisms are needed to prevent LLM hallucinations in optimization contexts.

\paragraph{Advanced Training and Refinement Paradigms.}

To enhance the reliability of LLM reasoning, recent work moves beyond simple pipelines to incorporate advanced training and refinement paradigms~\cite{fang2025comprehensive,xi2025survey}. One major direction is reinforcement learning, which is applied with varying feedback granularity. Outcome-driven methods like Solver-Informed RL (SIRL)~\citep{chen2025solverinformedrlgroundinglarge} use a solver’s final verdict for precise rewards but suffer from a severe credit assignment problem, as a correct final model might be reached via flawed intermediate steps \citep{chen2025solverinformedrlgroundinglarge}. In contrast, process-driven methods using Process Reward Models (PRMs) provide denser, step-by-step feedback but can be myopic, rewarding locally correct steps that are globally inconsistent \citep{dai2025processsupervisionguidedpolicyoptimization,zhu-etal-2025-retrieval}. To implement such reward schemes, preference-based training objectives like Direct Preference Optimization (DPO) emerge as a stable and effective alternative to complex RLHF pipelines \citep{zheng2025cold}. Complementing these training methods are iterative refinement and search strategies. These techniques embed LLMs into structured search frameworks like beam search \citep{Wang2024} and Monte Carlo Tree Search \citep{Zheng2025, Liu2024b}, or employ evolutionary loops with self-reflection to iteratively improve generated solutions \citep{Ye2024}. While these approaches make final outputs more reliable by exploring a wider solution space, they often operate at the output level rather than being deeply integrated into the model's core training process. Thus, a trade-off remains: outcome-based RL lacks intermediate supervision, stepwise PRMs lack global coherence, and search methods are often applied post-hoc. To bridge this gap, discriminative methods based on both final outcome correctness and stepwise feedback can penalize incorrect reasoning steps while maintaining global coherence, effectively combining the strengths of both process- and outcome-based approaches~\citep{chen2025stepwiseguidedpolicyoptimization}.

The research community actively builds end-to-end OR modeling systems that merge symbolic rigor with learned reasoning. This effort is supported by a growing number of benchmarks \citep{Wang2024, huang2024mamo, tang2024orlm, yang2025optibenchmeetsresocraticmeasure, xiao2025survey}. StepORLM is designed to address these gaps by embedding process-level supervision and iterative refinement, ensuring each modeling step is verified. Its co-evolutionary framework combines a generative PRM for nuanced step-level feedback with final-solver rewards, mitigating the credit assignment problem via a weighted DPO objective. By having its policy and generative PRM co-evolve, StepORLM learns to self-verify internally, leading to more robust and error-resistant modeling behavior that advances the goal of reliable and scalable OR reasoning with LLMs.

\section{Methodology}
\label{sec:method}

\subsection{Overview of StepORLM}
\begin{figure}[t]
  \centering
  \includegraphics[width=\linewidth]{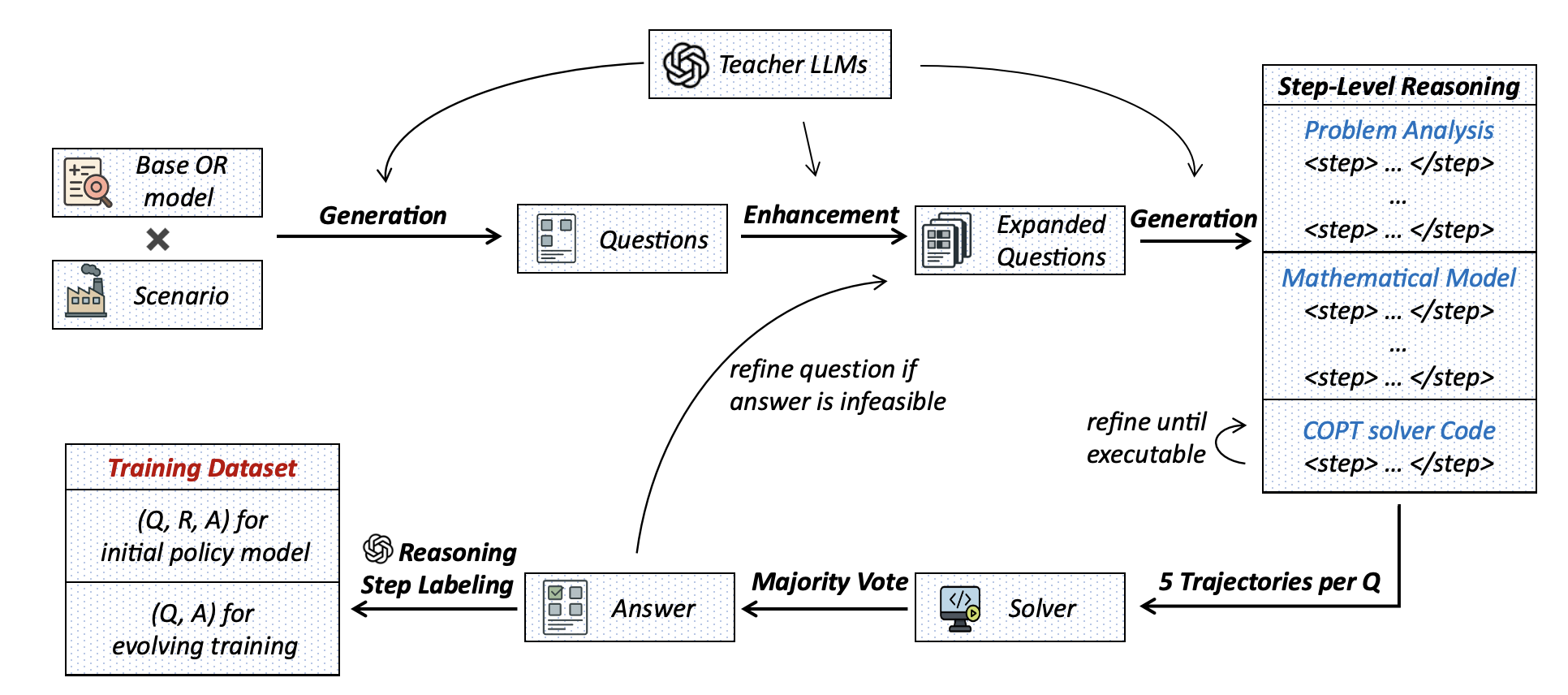}
  \vspace{-20pt}
  \caption{The illustration of data synthesis pipeline for the warm-up stage. 
  A teacher LLM generates diverse OR problems and their step-by-step reasoning solutions. Each solution is rigorously validated and filtered by an external solver, with an automated refinement loop to ensure the quality. 
  This verified corpus is then used to train the initial policy model via SFT.}
  \label{fig:data_synthesis_framework}
\end{figure}

\begin{figure}[t]
  \centering
  \includegraphics[width=\linewidth]{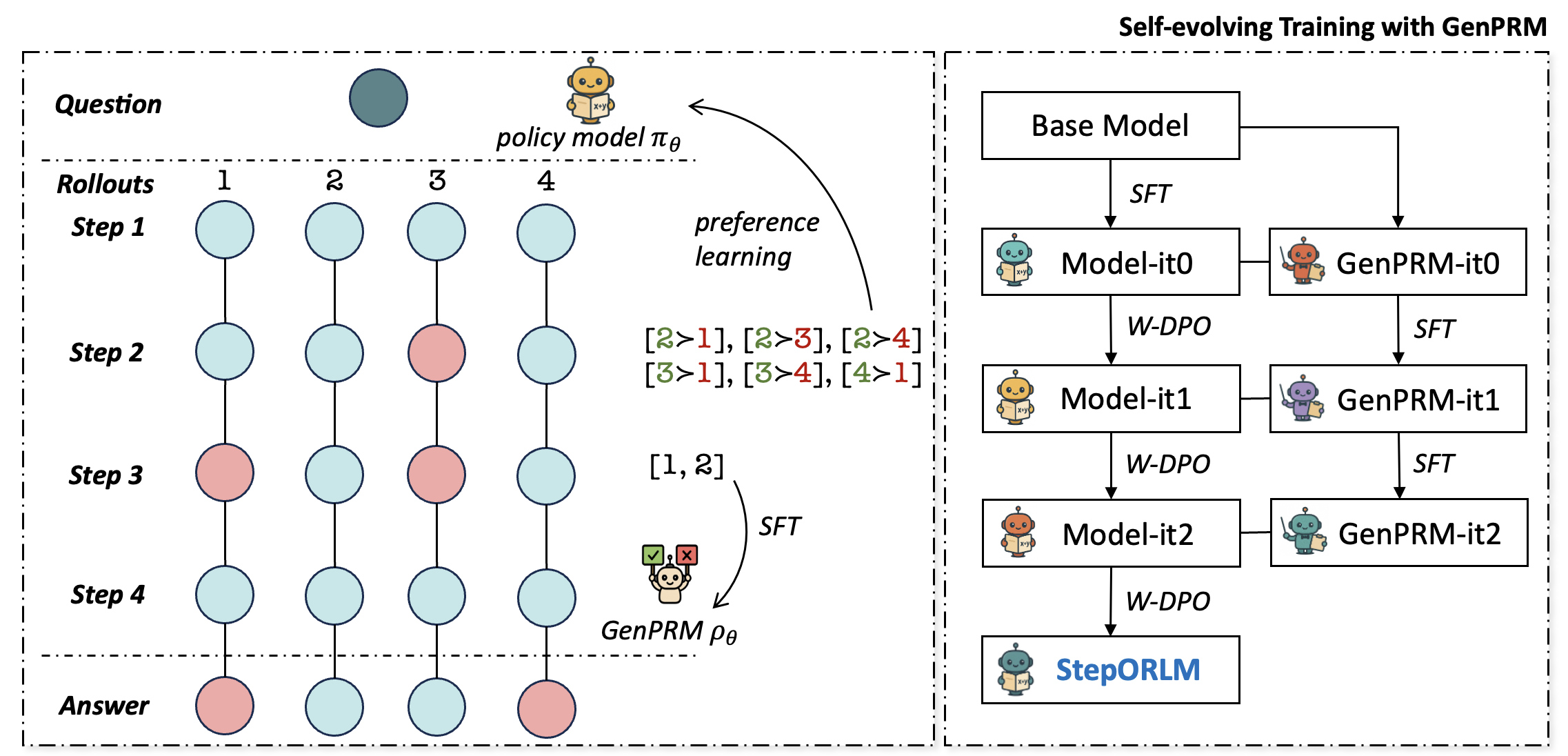}
  \vspace{-15pt}
  \caption{The co-evolutionary loop of \model{}. At each iteration, the policy model $\pi_{\theta}$ generates multiple trajectories. The feedback from both the external solver (outcome) and the GenPRM $\rho_{\theta}$ (process) is used to create training data that simultaneously refines the policy via W-DPO and improves the GenPRM via SFT, fostering reciprocal improvement.}
  \label{fig:self-evolve}
\end{figure}

We train \model{} using a two-stage, self-evolving framework designed to instill robust, process-sound reasoning for operations research problems. 
As illustrated in Figure~\ref{fig:data_synthesis_framework} and Figure~\ref{fig:self-evolve}, the framework is composed of two main stages:
\begin{itemize}[leftmargin=10pt]
    \item \textbf{Stage 1: Warm-Up for Initial Policy}. As shown in Figure~\ref{fig:data_synthesis_framework}, we bootstrap the training process by creating a high-quality, solver-verified dataset, which consists of OR problems and step-by-step solutions. 
    This corpus is used to train an initial policy $\pi_0$ via Supervised Fine-Tuning (SFT).
    \item \textbf{Stage 2: Iterative Co-Evolution for Policy and GenPRM}. As shown in Figure~\ref{fig:self-evolve}, starting with the initial policy $\pi_0$ and a base LLM as the critic $\rho_{\theta_0}$, we commence a self-evolving training loop. The policy model and the generative process reward model (GenPRM) co-evolve, with the policy generating increasingly better solutions and the GenPRM becoming a more discerning critic.
\end{itemize}

\subsection{Warm-Up for Initial Policy}
\label{sec:warmup}
The goal of the warm-up stage is to produce a strong initial policy model capable of generating structured, multi-step solutions to OR problems. We achieve this through a scalable data synthesis pipeline, shown in Figure~\ref{fig:data_synthesis_framework}.

We begin with a large pool of base OR problem templates and diverse industry scenarios. A powerful teacher LLM (GPT-4o in this paper) first pairs compatible templates and scenarios to draft a natural-language question $Q$. To enrich the data, each question is augmented through paraphrasing, unit conversion, and controlled parameter scaling. The teacher model then generates a complete, multi-step reasoning trajectory for each question $Q$, mirroring the workflow of a human expert from problem analysis to executable solver code. See Appendix~\ref{appendix:initial_trajectory} for more details.

To ensure data quality, every generated solution undergoes a deterministic validation process: we execute the code, inspect the solver status, and verify the objective value. 
Trajectories that produce errors would trigger an automatic self-refinement loop, prompting the teacher LLM to revise the formulation or code until validation succeeds or the retry budget is exhausted. For questions with multiple valid solutions, we select the one achieving the best objective value. This rigorous process yields a high-quality dataset of verified $(Q, R)$ pairs, i.e., Question (Q) and Reasoning (R). We use the constructed dataset to train the initial policy model $\pi_0$ via Supervised Fine-Tuning (SFT).

\subsection{Iterative Co-Evolution for Policy and GenPRM}
\label{sec:iter_evolution}
Following the warm-up stage, the iterative co-evolution stage begins. As illustrated in Figure~\ref{fig:self-evolve}, this self-evolving loop simultaneously enhances the policy model $\pi_\theta$ and the generative process reward model (GenPRM) $\rho_\phi$. Each iteration involves three key steps: 1) trajectory collection and evaluation, 2) policy alignment, and 3) GenPRM refinement.

\subsubsection{Trajectory Collection and Dual-Source Evaluation}

Our training data is collected in the warm-up stage.
For each problem in the training set, the current policy $\pi_{\theta}$ generates a set of $k$ candidate solution trajectories. These trajectories are then put through a dual-source evaluation:
\begin{itemize}[leftmargin=15pt]
    \item \textbf{Outcome Verification}. The final code of each trajectory is executed, and an external solver provides a definitive, ground-truth label of success or failure.
    \item \textbf{Process Evaluation}. The GenPRM $\rho_{\theta}$ holistically assesses each complete trajectory and provides step-by-step correctness scores, capturing the quality of the reasoning process. 
    Note that, different from conventional discriminative process supervision, our GenPRM is required to first perform chain-of-thought reasoning over the OR solution and then generate a holistic, post-hoc evaluation of the entire reasoning trajectory.
\end{itemize}
After assessing the generated trajectories, we use the dual-feedback signals to further refine both the policy and GenPRM. The improved models would then start another round of self-evolving iteration. 
Next, we introduce the policy alignment and GenPRM refinement at each iteration.

\subsubsection{Policy Alignment via W-DPO}
The dual-source feedback is first used to align the policy model. We distill the trajectory evaluation results into preference pairs $(\tau_w, \tau_l, w)$, where $\tau_w$ is the winning trajectory and $\tau_l$ is the losing one. 
The selection logic, as detailed in Algorithm~\ref{alg:preference}, prioritizes solver-verified solutions. 
If solver outcomes are identical for a pair of trajectories (e.g., both pass or both fail), then the trajectory with a higher ratio of correct steps, as judged by the GenPRM, is preferred.

Next, we design a Weighted Direct Preference Optimization (W-DPO) objective for policy training. The weight $w$ is defined to quantify the quality gap between two trajectories: solver-distinguished pairs receive a fixed, high weight (i.e., $w=1.0$), while others are weighted proportionally to the difference in their process quality scores (i.e., ratio of correct steps). 
The final objective focuses the model training on the most informative pairs weighted by $w$:
\begin{equation}
    \mathcal{L}_{\mathrm{W\text{-}DPO}}(\theta)
    \;=\;
    -\,\mathbb{E}_{(x, \tau_w, \tau_l)} 
    \Big[ w_{(\tau_w, \tau_l)} \cdot \log \sigma \!\big( 
        \beta \big( \log \pi_\theta(\tau_w \mid x) - \log \pi_\theta(\tau_l \mid x) \big) 
    \big) \Big],
    \label{eq:wdpo}
\end{equation}
where $w_{(\tau_w, \tau_l)}$ is the weight for trajectory pair $(\tau_w, \tau_l)$, $\pi_\theta(\tau \mid x)$ is the log-likelihood of generating trajectory $\tau$ given the input question $x$, and $\beta$ is a scaling factor.

\begin{algorithm}[t]
\caption{Preference Pair Construction}
\label{alg:preference}
\begin{algorithmic}[1]
    \State \textbf{Input:} Two trajectories $\tau_a, \tau_b$.
    \State \textbf{Output:} A tuple $(\tau_w, \tau_l, w)$.
    
    \State \textcolor{blue}{\textit{// Rule 1: A solver-verified trajectory is always preferred.}}
    
    \If{$\tau_a$ has a correct result and $\tau_b$ does not}
        \State \Return $(\tau_a, \tau_b, 1.0)$  
    \ElsIf{$\tau_b$ has a correct result and $\tau_a$ does not} 
        \State \Return $(\tau_b, \tau_a, 1.0)$
    \EndIf
    
    \State \textcolor{blue}{\textit{// Rule 2: Otherwise, prefer the trajectory with a higher ratio of correct steps.}}
    
    \State Let $\text{diff} \leftarrow \text{CorrectStepRatio}(\tau_a) - \text{CorrectStepRatio}(\tau_b)$
    \If{$\text{diff} > 0$}
        \State \Return $(\tau_a, \tau_b, \text{diff})$
    \ElsIf{$\text{diff} < 0$}
        \State \Return $(\tau_b, \tau_a, -\text{diff})$
    \Else
        \State \Return None
    \EndIf
\end{algorithmic}
\end{algorithm}

\subsubsection{GenPRM Refinement via SFT}
The same trajectory evaluation data from the current iteration is then used to refine the critic, i.e., GenPRM. 
Specifically, we extract all step-level correctness labels derived from the solver-verified trajectories to form a new, high-quality dataset. 
This dataset is used to continue the supervised fine-tuning of GenPRM $\rho_{\theta}$ with process supervision signals.

We can see that such a process creates a positive feedback loop: as the policy improves, it generates more diverse and accurate reasoning paths, providing higher-quality training data for the GenPRM. In turn, a more capable GenPRM provides a more precise and reliable reward signal, further accelerating the policy alignment. This symbiotic refinement is crucial for surpassing the performance of the initial SFT model, which is thus referred to as the self-evolving framework.

\section{Experiments}

\subsection{Experiment Setup}\label{sec:setup}

\paragraph{Benchmarks.}
We conduct our evaluation on a diverse set of six public benchmarks widely used in the OR community: NL4Opt~\citep{Ramamonjison2023}, MAMO (split into EasyLP and ComplexLP)\citep{huang2024mamo}, NLP4LP\citep{AhmadiTeshnizi2024}, ComplexOR\cite{xiao2024chainofexperts}, IndustryOR\citep{tang2024orlm} and ReSocratic\citep{yang2025optibenchmeetsresocraticmeasure}. More details about the six benchmarks are provided in appendix~\ref{appendix:benchmarks}.

\paragraph{Baselines.}
To ensure a thorough comparison, we benchmark against a comprehensive set of baselines in three categories:
\begin{itemize}[leftmargin=15pt]
\item \textbf{Zero-shot Generalist LLMs:} Powerful, large-scale models without specific fine-tuning, including \texttt{GPT-4o} \citep{openai2024gpt4o}, \texttt{DeepSeek-V3} \citep{deepseekai2025deepseekv3technicalreport}, \texttt{Qwen3-32B} \citep{yang2025qwen3technicalreport}, and \texttt{Qwen2.5-72B-Instruct} \citep{qwen2025qwen25technicalreport}.
\item \textbf{Specialized Fine-tuned LLMs:} Models specifically trained on OR solving tasks, including \texttt{ORLM} (8B) \citep{tang2024orlm}, \texttt{LLMOPT} (14B) \citep{Jiang2025a} and \texttt{OptMATH} (32B) \citep{lu2025optmathscalablebidirectionaldata}. We try our best to re-evaluate all publicly accessible models. 
For those not open-sourced, we report their original performance according to corresponding published papers. 
Note that due to our use of a cleaned dataset~\citep{xiao2025survey}, these original scores serve as a reference and may not be directly comparable.
\item \textbf{Agentic Methods:} Multi-step reasoning frameworks that orchestrate LLMs to solve problems, including \texttt{OptiMUS-v0.3} \citep{optimus0.3}, Chain-of-Thought (\texttt{CoT}) \citep{wei2022chainofthought}, Chain-of-Experts (\texttt{CoE}) \citep{xiao2024chainofexperts}, and \texttt{CAFA} \citep{deng2024cafa}. To ensure a fair and powerful comparison, all agentic methods are implemented using \texttt{GPT-4o} as their backbone model.
\end{itemize}

\paragraph{Implementation Details}
We develop our \texttt{StepORLM} models by fine-tuning the publicly available Qwen3-8B \citep{yang2025qwen3technicalreport}. 
Training and inference are conducted on a single node with 8 × NVIDIA H100 80 GB GPUs. We used a synthesized dataset of 50K questions for subsequent training. The model was trained using the AdamW optimizer with a learning rate of 7e-6 and weight decay of 0.1. The maximum sequence length was set to 8192, and we introduced special tokens \texttt{<step>} and \texttt{</step>} to enhance reasoning. A batch size of 64 was used, and training proceeded for 3 epochs. During the self-evolving sampling phase, we generated 4 trajectories for each response to be used for subsequent preference learning. For W-DPO, we set the hyperparameter $\beta$ to 0.1. To ensure optimal performance and a fair comparison across different frameworks, we selected solvers best suited to each category of model. For general-purpose LLMs and all agentic methods, we used Gurobi due to its broad compatibility. For the fine-tuned models, we used their respectively adapted solvers: both our \texttt{StepORLM} and \texttt{ORLM} employed the COPT solver, whereas \texttt{LLMOPT} is evaluated using Pyomo. 

\begin{table*}[t]
\footnotesize
\centering
\caption{The overall performance of \model{} and baselines with Pass@1 accuracy (\%) on six OR benchmarks. 
\texttt{StepORLM+GenPRM} indicates using the GenPRM as process verifier to enable the inference scaling of \model{}. 
Scores cited from original publications are marked with the symbol (*), while missing entries are denoted with (-). 
Best results are highlighted in \textbf{bold} and the second-highest values are \underline{underlined}.}
\label{table:main_results}
\setlength{\tabcolsep}{2pt}
\adjustbox{max width=1.0\textwidth}{
\begin{tabularx}{\textwidth}{l*{9}{>{\centering\arraybackslash}X}}
\toprule
Model & Params & NL4OPT & \multicolumn{2}{c}{MAMO} & NLP4LP & CompOR & IndOR & ReSocratic & Avg. \\ 
\cmidrule(lr){4-5}
 & & & EasyLP & ComplexLP & & & & & \\ 
\midrule
\rowcolor{gray!15} 
\multicolumn{10}{c}{\bf \it Zero-shot LLMs} \\
\texttt{GPT-4o} & Closed & 61.2 & 70.3 & 57.7 & 73.6 & 42.9 & 38.1 & 48.4 & 56.0 \\
\texttt{DeepSeek‑V3} & 671B & 79.8 & 95.2 & 53.2 & 92.1 & 55.6 & 66.7 & 85.1 & 75.4 \\
\texttt{Qwen3-32B} & 32B & 77.5 & 92.3 & 46.9 & 93.8 & 50.0 & 61.9 & 85.1 & 72.5 \\
\texttt{Qwen2.5-72B-Inst} & 72B & 78.9 & 95.8 & 44.1 & 88.2 & 50.0 & 57.1 & 81.1 & 70.7 \\
\midrule
\midrule
\rowcolor{gray!15} 
\multicolumn{10}{c}{\bf \it Fine-tuned LLMs} \\ 
\texttt{ORLM} & 8B & 73.8 & \underline{90.4} & \underline{59.5} & \underline{76.4} & \underline{50.0} & \underline{42.9} & \underline{61.8} & \underline{65.0} \\
\texttt{LLMOPT} (origin) & 14B & 80.3* & 89.5* & 44.1* & 73.4* & 35.3* & 29.0* & 53.8* & 57.9* \\
\texttt{LLMOPT} (reproduce) & 14B & 49.3 & 36.3 & 25.2 & 43.3 & 16.7 & 40.5 & 39.5 & 35.8 \\
\texttt{OptMATH} (origin) & 32B & \underline{95.9*} & 89.9* & 54.1* & - & - & - & - & - \\
\rowcolor{purple!15}
\texttt{StepORLM} & 8B & \textbf{96.7} & \textbf{97.6} & \textbf{77.5} & \textbf{97.2} & \textbf{50.0} & \textbf{52.4} & \textbf{81.9} & \textbf{79.0} \\
\midrule
\midrule
\rowcolor{gray!15} 
\multicolumn{10}{c}{\bf \it Agentic Methods} \\
\texttt{OptiMUS-v0.3} & Closed & \underline{76.2} & 78.0 & 46.8 & \underline{88.8} & 46.8 & \underline{45.2} & \underline{87.6} & \underline{67.1} \\
\texttt{CoT} & Closed & 62.2 & 49.5 & 42.3 & 74.7 & 39.2 & 40.5 & 43.6 & 50.3 \\
\texttt{CoE} & Closed & 66.7 & \underline{94.4} & \underline{50.6} & 87.4 & \underline{57.1} & 31.2 & 71.2 & 65.5 \\
\texttt{CAFA} & Closed & 68.1 & 71.2 & 44.5 & 50.0 & 46.4 & 41.1 & 40.1 & 51.6 \\
\rowcolor{purple!15}
\texttt{StepORLM+GenPRM} & 8B+8B & \textbf{97.2} & \textbf{97.8} & \textbf{87.4} & \textbf{98.9} & \textbf{61.1} & \textbf{61.9} & \textbf{94.6} & \textbf{85.6} \\
\bottomrule
\end{tabularx}
}
\par
\scriptsize
\textbf{Abbreviations}: CompOR: ComplexOR, IndOR: IndustryOR, Avg: Macro-Average, Qwen2.5-72B-Inst: Qwen2.5-72B-Instruct.
\end{table*}

\subsection{Main Results}\label{sec:main-result}

We report the main results in Table~\ref{table:main_results}. Our \model{} establishes a new state-of-the-art in OR problem-solving tasks, significantly outperforming both vastly larger generalist/agentic models and specialized LLMs.
As an 8B-parameter model, \model{} outperforms much larger, state-of-the-art generalist models, such as \texttt{DeepSeek-V3} (671B) and \texttt{Qwen2.5-72B-Instruct} (72B), and even those agentic methods based on \texttt{GPT-4o}.
This result underscores the profound effectiveness of specialized, process-supervised training, enabling a compact model to achieve more reliable and accurate OR modeling capabilities than its far larger counterparts.

Moreover, leveraging the co-evolved GenPRM as an inference-time process verifier (i.e., \texttt{\model{}+GenPRM}) further pushes the performance to an average Pass@1 accuracy of 85.6\%. 
Such a configuration excels especially on the most challenging benchmarks, with remarkable accuracy gains of 9.9\% on ComplexOR and 9.5\% on IndustryOR compared with the basic \model{}.
This highlights the dual benefit of our co-evolutionary framework: the policy model learns to generate high-quality solutions, while the GenPRM learns to effectively identify and select the best one. Furthermore, StepORLM excels at avoiding reward hacking, a point we illustrate with a case study in Appendix~\ref{appendix:case_study}.

\begin{figure}[t]
    \centering
    \includegraphics[width=1\linewidth]{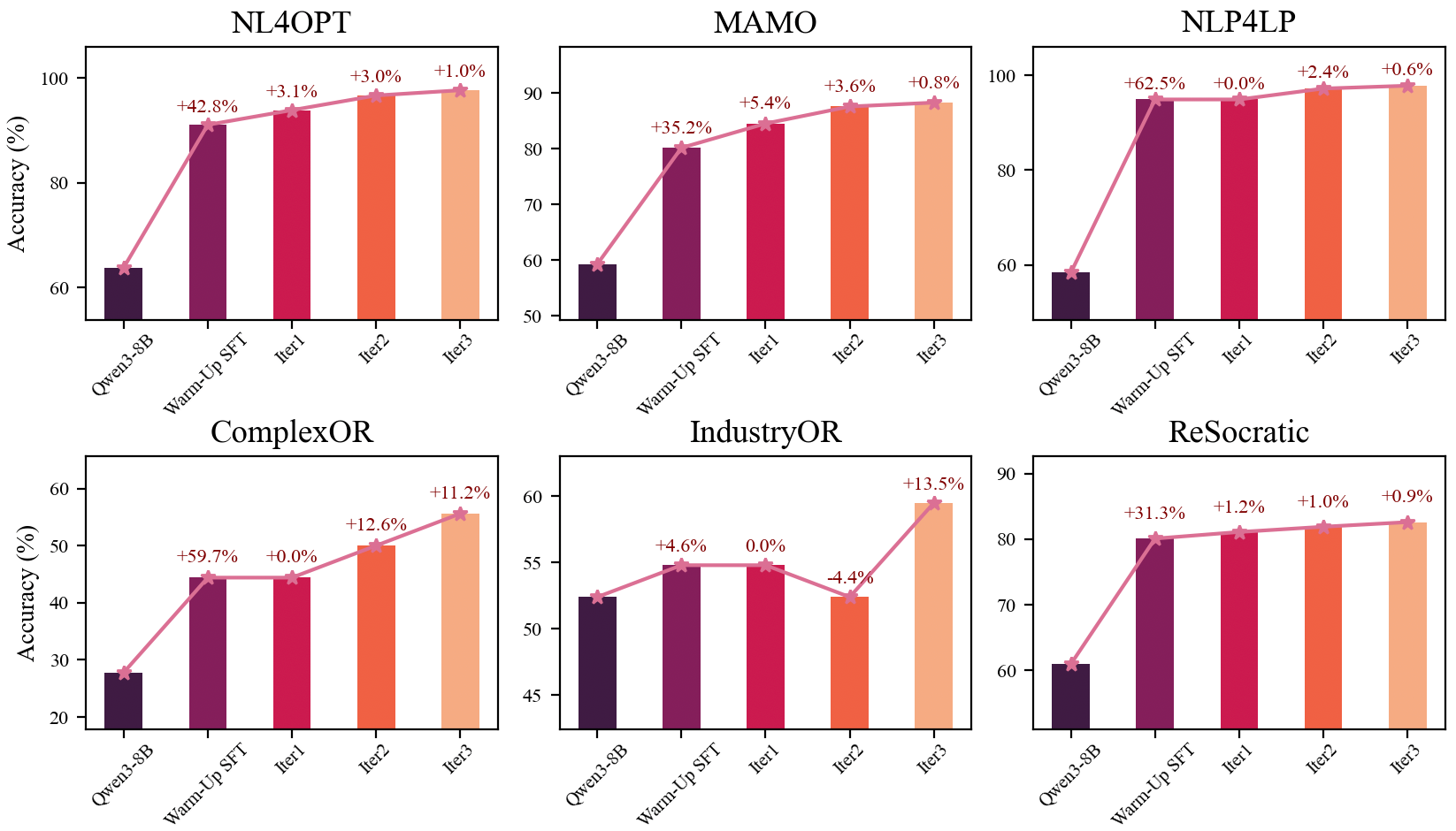}
    \vspace{-20pt}
    \caption{The analysis on the self-evolving process by tracking the performance (Pass@1 accuracy) at each training iteration. 
    The relative improvement of current iteration over the previous one is demonstrated on the corresponding bar.}
    \label{fig:self-evolving}
\end{figure}
\subsection{Analysis of Self-Evolving Process}\label{sec:self-evolving-process}

We conduct in-depth analysis on the self-evolving process by tracking the performance at each training iteration, including the warm-up and co-evolving stages.
The results are reported in Figure~\ref{fig:self-evolving}, from which we can obtain the following observations:
\begin{itemize}[leftmargin=15pt]
    \item The warm-up supervised fine-tuning (SFT) provides a massive foundational performance lift across all benchmarks (e.g., 62.5\% accuracy improvement on NLP4LP).
    \item The subsequent self-evolution iterations consistently build upon the prior models, delivering incremental but crucial accuracy improvements. This confirms that the model progressively refines its capabilities through our self-evolving loop, rather than stagnating after initial fine-tuning.
    \item The performance trend is notably non-monotonic on the IndustryOR dataset, where the performance first declines and finally largely increases at the third iteration. We attribute this to the small size of testing set and the challenging questions in IndustryOR. 
    Our case-level inspection reveals a potential progression. Early iterations primarily rectify structural modeling errors flagged by the PRM, whereas later iterations concentrate on code-level issues (e.g., index-out-of-bounds errors), which finally leads to accuracy improvements.
\end{itemize}

\subsection{Inference Scaling with GenPRM}\label{sec:testtime}

In this section, we investigate the generalization capability of our co-evolved GenPRM as the process verifier for inference scaling. 
We apply it to our final \model{} and another open-sourced \texttt{ORLM} baseline, and compare GenPRM's effectiveness against other inference scaling strategies including majority voting, solver execution, and discriminative PRM.

As shown in Table~\ref{table:inference scaling}, our self-evolved GenPRM is generally the most effective inference scaling strategy, achieving the highest accuracy on average for both \model{} and \texttt{ORLM}. 
When paired with \texttt{ORLM}, our GenPRM boosts its average performance by 10\%, demonstrating its powerful generalization.
It learns fundamental, model-agnostic principles of valid OR reasoning, allowing it to function as a universal verifier that significantly improves the performance of other models, not just its co-evolved policy.

\begin{table*}[t]
    \footnotesize
    \centering
    \caption{Performance comparison of \model{} under different inference scaling strategies.
     }
    \label{table:inference scaling}
    \setlength{\tabcolsep}{2pt}
    \vskip 0.15in
    \adjustbox{max width=1.0\textwidth}{
    \begin{tabularx}{\textwidth}{l*{8}{>{\centering\arraybackslash}X}}
    \toprule
    Model & NL4OPT & \multicolumn{2}{c}{MAMO} & NLP4LP & CompOR & IndOR & ReSocratic & Avg. \\ 
    \cmidrule(lr){3-4}
     & & EasyLP & ComplexLP & & & & & \\ 
    \midrule
    \rowcolor{gray!15} 
    \multicolumn{9}{c}{\bf \it \texttt{\textbf{StepORLM as Policy Model}}} \\ 
    \texttt{StepORLM} & \underline{97.7} & 97.2 & 79.3 & \underline{97.8} & \underline{55.6} & 59.5 & 82.6 & 81.4 \\ 
    \quad +\,Major Vote & 97.2 & 97.6 & 81.1 & 96.6 & \textbf{61.1} & \underline{61.9} & 89.3 & 83.5 \\
    \quad +\,Solver Exec & \underline{97.7} & \textbf{98.4} & 81.1 & 96.1 & \textbf{61.1} & \textbf{66.7} & 90.3 & \underline{84.5} \\
    \quad +\,Discriminative PRM & 97.2 & 97.2 & 81.1 & 97.2 & 55.6 & 59.5 & 87.8 & 82.2 \\
    \quad +\,GenPRM (initial) & \textbf{97.8} & 97.6 & \underline{82.8} & 97.2 & \underline{55.6} & 58.5 & \underline{93.1} & 83.2 \\
    \rowcolor{purple!15}
    \quad +\,GenPRM (final) & 97.2 & \underline{97.8} & \textbf{87.4} & \textbf{98.9} & \textbf{61.1} & \underline{61.9} & \textbf{94.6} & \textbf{85.6} \\
    \midrule
    \midrule
    \rowcolor{gray!15} 
    \multicolumn{9}{c}{\bf \it \texttt{\textbf{ORLM as Policy Model}}} \\ 
    \texttt{ORLM} & 73.8 & 90.4 & 59.5 & 76.4 & \textbf{50.0} & 42.9 & 61.8 & 65.0 \\
    \quad +\,Major Vote & 78.7 & 88.4 & 50.5 & 78.7 & \underline{44.4} & 47.6 & 73.0 & 65.9 \\
    \quad +\,Solver Exec & 82.2 & 88.6 & \underline{63.1} & 79.8 & \underline{44.4} & 52.4 & \underline{78.9} & 69.9 \\
    \quad +\,Discriminative PRM & 75.1 & \textbf{91.7} & \underline{63.1} & 82.0 & \textbf{50.0} & \underline{54.8} & 74.7 & \underline{70.2} \\
    \quad +\,GenPRM (initial) & \underline{87.3} & 90.6 & 55.0 & \underline{90.4} & \underline{44.4} & 47.6 & 65.5 & 68.7 \\
    \rowcolor{purple!15}
    \quad +\,GenPRM (final) & \textbf{91.5} & \underline{91.0} & \textbf{64.9} & \textbf{91.0} & \textbf{50.0} & \textbf{57.1} & \textbf{79.4} & \textbf{75.0} \\
    \bottomrule
    \end{tabularx}
    }
    \par
    \scriptsize
    \vskip -0.1in
\end{table*}

\begin{table*}[t]
    \footnotesize
    \centering
    \caption{Ablation study of \model{} on Pass@1 accuracy (\%).
    The best results are given in \textbf{bold}, and the second-best values are \underline{underlined}.}
    \label{table:ablation}
    \setlength{\tabcolsep}{2pt}
    \vskip 0.15in
    \adjustbox{max width=1.0\textwidth}{
    \begin{tabularx}{\textwidth}{l*{8}{>{\centering\arraybackslash}X}}
    \toprule
    Model & NL4OPT & \multicolumn{2}{c}{MAMO} & NLP4LP & CompOR & IndOR & ReSocratic & Avg. \\ 
    \cmidrule(lr){3-4}
     & & EasyLP & ComplexLP & & & & & \\ 
    \midrule
    \texttt{StepORLM} & \textbf{97.7} & \textbf{97.2} & \textbf{79.3} & \textbf{97.8} & \textbf{55.6} & \textbf{59.5} & \textbf{82.6} & \textbf{81.4} \\
    w/o SFT & 71.4 & 88.3 & 55.9 & 72.5 & 38.9 & 50.0 & 75.2 & 64.6 \\
    w/o Self-evolution & 96.7 & \textbf{97.2} & 67.6 & 94.9 & 44.4 & 52.4 & 81.4 & 76.4 \\
    w/o GenPRM Evolution & 94.4 & \underline{96.3} & 68.5 & \textbf{97.8} & \underline{50.0} & \underline{54.8} & \underline{82.4} & 77.7 \\
    w/o W-DPO & \underline{97.2} & 95.8 & \underline{73.9} & \underline{96.1} & \underline{50.0} & 52.4 & 80.6 & \underline{78.0} \\
    \bottomrule
    \end{tabularx}
    }
    \par
    \scriptsize
    \vskip -0.1in
\end{table*}
\subsection{Ablation Study}\label{sec:ablation}

To study the impact of each component in our proposed self-evolving framework, we conduct experiments based on the following variants: 
(1) w/o SFT: skipping the warm-up supervised fine-tuning stage; (2) w/o self-evolution: directly sampling a single reasoning trajectory after SFT and applying DPO without iterative refinement; (3) w/o GenPRM Evolution: using the initial GenPRM with frozen parameters for all subsequent iterations; (4) w/o W-DPO: replacing the W-DPO with standard DPO during the self-evolving training phase. 
The results are shown in Table~\ref{table:ablation}.

Removing the warm-up SFT (w/o SFT) causes the largest performance drop, confirming the necessity of a strong initial policy. 
Disabling the iterative loop (w/o Self-evolution) prevents the model from refining its initial capabilities, leading to lower scores on challenging benchmarks like ComplexOR. 
Freezing the GenPRM after initialization (w/o GenPRM Evolution) also degrades performance, validating the importance of the co-evolutionary dynamic between the policy and the critic. 
Finally, replacing our W-DPO objective with standard DPO (w/o W-DPO) results in a noticeable performance decrease, indicating that our trajectory-level quality weighting provides a more effective learning signal between different trajectory pairs. Together, these results demonstrate that the synergy among all components drives the success of the \model{} framework.

\section{Conclusion}

In this paper, we design \textbf{\model{}}, a novel self-evolving framework to overcome critical limitations in training large language models for operations research. Our approach addresses the credit assignment problem of outcome-based rewards and the myopia of conventional process supervision through a co-evolutionary loop between a policy model and a generative process reward model (GenPRM).
\model{} integrates definitive solver-based outcome verification with holistic, generative process feedback. 
Such a dual-feedback mechanism ensures the process-sound and outcome-correct reasoning capabilities of the policy model to solve complex OR problems.
Our 8B-parameter \model{} achieves the SOTA performance across six benchmarks, significantly outperforming much larger generalist models, agentic systems, and specialized baselines. 
Furthermore, the co-evolved GenPRM is able to act as a powerful, universal inference-time process verifier, substantially enhancing the performance of our model and other LLMs for OR problem-solving tasks.

\bibliographystyle{iclr2026_conference}
\bibliography{ref}

\newpage
\appendix
\onecolumn

\section{Benchmarks}\label{appendix:benchmarks}
     A mass of datasets and methodologies for dataset construction have been proposed to improve the capacity of LLM to address OR problems, with these datasets becoming increasingly intricate and realistic\citep{AhmadiTeshnizi2024, yang2025optibenchmeetsresocraticmeasure}. A recent survey by \citep{xiao2025survey} provides a comprehensive and timely review of benchmark datasets for optimization problems described in natural language. The study identified prevalent issues within the datasets, including logical fallacies, ambiguous problem definitions and incorrect ground-truth answers, and also corrected a portion of the errors identified. Based on that, our work builds upon their initial efforts by conducting further manual verification and cleaning of these widely-used benchmarks. Subsequently, the data is processed into a unified format, thereby establishing a more robust and standardized foundation for subsequent research and model assessment.

    \subsection{NL4Opt}
    \label{sec:nl4opt}
    The NL4Opt dataset was introduced by Ramamonjison et al. as part of the NeurIPS 2022 NL4Opt competition \citep{ramamonjison2022augmenting,Ramamonjison2023}. It consists of approximately 1,100 annotated word problems describing linear programming (LP) scenarios, with the primary purpose of bridging natural language descriptions and formal optimization models. In a survey, \citeauthor{xiao2025survey} reported that NL4Opt has a complexity score of 5.59 and an original error rate of at least 26.4\%. Each problem presents a self-contained, real-world scenario (e.g., resource allocation, scheduling) with annotations identifying key optimization elements. The official shared task for the dataset is divided into two subtasks: (1) recognizing and labeling problem entities and (2) generating a corresponding mathematical model. The original dataset provides 713 training, 99 validation, and 289 test instances. For our evaluation, we selected 213 validated problems from this benchmark.

    \subsection{MAMO}
    \label{sec:mamo}
    The MAMO dataset was introduced by Huang et al. (2024) as a benchmark for evaluating large language models on complex mathematical modeling tasks, with a specific focus on the formulation process itself rather than just solution accuracy \citep{huang2024mamo}. The dataset concentrates on linear programming (LP) and mixed-integer linear programming (MILP) problems, excluding nonlinear or differential equation modeling. MAMO is divided into two main components: Easy LP (652 instances) and Complex LP (211 instances). According to an analysis by \citeauthor{xiao2025survey}, EasyLP has a complexity of 7.12 with an error rate of at least 8.13\%, while ComplexLP is significantly more difficult, with a complexity of 13.35 and an error rate of at least 23.7\%. By delegating the solving step to an OR solver, MAMO focuses specifically on whether an LLM can accurately translate a detailed natural language description into a correct mathematical formulation. From this benchmark, we selected 545 easy and 111 complex instances for our study.

    \subsection{NLP4LP}
    \label{sec:nlp4lp}
    The NLP4LP benchmark, introduced by AhmadiTeshnizi et al. (2023) in their work on the OptiMUS system \citep{AhmadiTeshnizi2024,optimus0.3}, evaluates LLM-based agents on translating natural language operations research problems into solver-ready code and mathematical models. The survey by \citeauthor{xiao2025survey} assigned it a complexity score of 5.58 and identified an error rate of at least 21.7\% in the original dataset. It consists of 269 human-authored LP and MILP problems, where each instance presents an optimization scenario from classical OR domains such as scheduling, knapsack allocation, and production planning. The benchmark provides a fine-grained testbed for assessing formulation accuracy. In our work, we chose to evaluate 178 verified instances from NLP4LP.

    \subsection{ComplexOR}
    \label{sec:complexor}
    The ComplexOR dataset was introduced by Xiao et al. as part of their Chain-of-Experts framework to stress-test the reasoning and modeling capabilities of LLMs on scenarios more challenging than those in earlier datasets \citep{xiao2024chainofexperts}. The analysis by \citeauthor{xiao2025survey} noted a complexity score of 5.98 and a significant error rate of at least 24.3\%. ComplexOR was curated from advanced OR case studies and classic difficult optimization problem classes, such as multi-step lot-sizing problems with setup costs, intricate scheduling tasks, and supply chain design problems. The problems are primarily large-scale or conceptually complex mixed-integer linear programs that require multi-step reasoning and often involve hierarchical or time-indexed structures. The dataset’s intended purpose is to facilitate the development and evaluation of advanced reasoning strategies for OR problem solving. Following our verification process, we selected 18 reliable problems from this dataset for evaluation.

    \subsection{IndustryOR}
    \label{sec:industryor}
    The IndustryOR dataset, introduced by Huang et al. (2024), is the first industrial operations research benchmark for large language models \citep{tang2024orlm}. It comprises 100 real-world optimization scenarios from sectors like manufacturing, logistics, finance, and energy. The study by \citeauthor{xiao2025survey} highlighted this dataset's high difficulty, reporting a complexity score of 14.06 and a substantial error rate of at least 54.0\%. Designed to evaluate LLMs on practical, domain-specific tasks, IndustryOR covers five OR categories—linear, integer, mixed-integer, nonlinear, and others—and is labeled by difficulty (Easy, Medium, Hard). For our experiments, we selected 42 validated problems from this benchmark.

    \subsection{ReSocratic}
    \label{sec:resocratic}
    The ReSocratic dataset is introduced by \citep{yang2025optibenchmeetsresocraticmeasure}, the name of which is actually a data synthesis method proposed by this work as well. This method synthesize the formatted optimization demonstration in a reverse manner first, and then back-translates it into a question. Through these intermediate reasoning steps, ReSocratic has a higher quality than previous methods. In this paper, we selected 403 validated problems from the original ReSocratic-29K dataset.

\section{Case Study}\label{appendix:case_study}
The following presentation will outline a comprehensive prototype of the failure case B, as referenced in Figure ~\ref{fig:opening-figure}. The sequence of events in the present case study is as follows: firstly, a sketch map of the problem is presented; secondly, a detailed problem description is given; and thirdly, the responses given by Qwen3 and our model StepORLM are outlined. As previously stated, the ensuing discourse will focus on the results obtained from the two models. The graph illustrates the problem and demonstrates that an error in the constraints step will not affect the optimal value.

\begin{figure}[htp]
    \centering
    \includegraphics[width=0.9\linewidth]{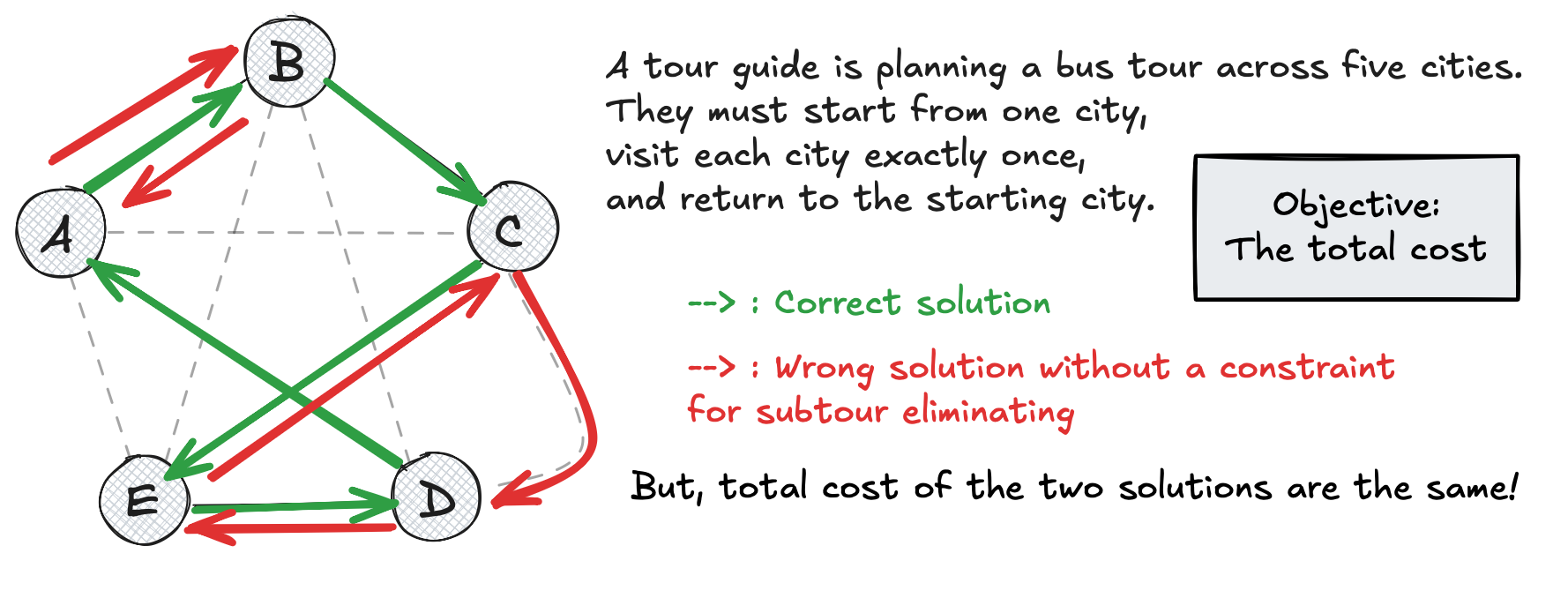}
    \caption{A simplified illustration of problem 74 in ComplexLP}
    \label{fig:enter-label}
\end{figure}

This is a problem with the identifier 74 from the ComplexLP dataset:
\begin{tcolorbox}[title=Problem Description, breakable, colback=orange!5]
    Consider a scenario where a tour guide is planning a bus tour across five cities, named E, F, G, H, and I. The tour must start and end in the same city, and each city should be visited exactly once. The objective is to minimize the total cost of the tour, which could be influenced by factors such as distance, tolls, and fuel expenses.\\

    Here are the travel costs between the cities:\\

    From City E, it costs 37 units to travel to F, 72 units to G, 66 units to H, and 33 units to I.
    Travelling from City F, the costs are 37 units to E, 26 units to G, 56 units to H, and a whopping 100 units to I.
    If the journey starts from City G, it takes 72 units to reach E, 26 units to F, 49 units to H, and 32 units to I.
    From City H, it costs 66 units to get to E, 56 units to F, 49 units to G, and a mere 13 units to I.
    Lastly, from City I, the travel costs are 33 units to E, 100 units to F, 32 units to G, and 13 units to H.\\

    Keeping these costs in mind, what is the minimum total travel cost for the tour guide to take the bus to each city exactly once and return to the starting city?
\end{tcolorbox}

And the following is our generate prompt for both Qwen3 and StepORLM.
\begin{tcolorbox}[title=Generate Prompt, breakable, colback=orange!5]
You are a helpful assistant with expertise in mathematical modeling, Python code and the COPT solver. When the User provides an optimization question, you will analyze it, build a detailed mathematical model, and provide the COPT code to solve it.\\

Your response should follow these steps:\\
1. Problem Description\\
2. Sets and Parameters\\
3. Decision Variables\\
4. Objective Function\\
5. Constraints\\
6. Mathematical Model\\
7. Nonlinear Relationships\\
8. Final Model\\
9. Python Code Using coptpy\\

please enclose the content within \textless step\textgreater ...\textless /step\textgreater.\\

Below is an optimization modeling question. Build a mathematical model and corresponding python code using coptpy that appropriately addresses the question: \\

PROBLEM DESCRIPTION HERE\\

Keeping these costs in mind, what is the minimum total travel cost for the tour guide to take the bus to each city exactly once and return to the starting city?\\

* Make sure to import necessary packages, such as 'import coptpy as cp' and 'from coptpy import COPT'.\\
* When you create a model make sure to use 'env = cp.Envr()' and 'model = env.createModel'\\
* When you add a variable, use 'vtype = COPT.'\\
* Do not name variables and constraints\\
* Use '.addConstr' or '.addConstrs' to add constraints. If you want to set 'lb' or 'ub' as infinity, please use 'lb=COPT.INFINITY' or 'ub=COPT.INFINITY' instead of 'cp.INFINITY'.\\
* When you set objective, you should use the 'model.setObjective' method and use 'COPT.MINIMIZE' or 'COPT.MAXIMIZE'.\\
* Do not use 'model.optimize()'.\\
* Make sure to use 'model.solve()' to solve the question.\\
* The code output statement is:\\
if model.status == COPT.OPTIMAL:
    solution = {var.getName(): var.X for var in model.getVars()}
    print('Just print the best obj:', model.ObjVal)
else:
    print('No Solution')\\
    
Think step by step.
\end{tcolorbox}

\subsection{StepORLM's response}
\begin{tcolorbox}[title=Inference, breakable, colback=orange!5]
$<step>$\\
\textbf{Problem Description:}\\
The tour guide needs to plan a bus tour that visits each of the five cities exactly once and returns to the starting city, minimizing the total travel cost. This problem is a classic example of the Traveling Salesman Problem (TSP), which is a well-known combinatorial optimization problem.\\
$</step>$\\
$<step>$\\
\textbf{Sets and Parameters:}\\
- \textbf{Cities}: E, F, G, H, I\\
- \textbf{Cost Matrix}: A matrix where each element \( c_{ij} \) represents the cost of traveling from city \( i \) to city \( j \).\\
$</step>$\\
$<step>$\\
\textbf{Decision Variables:}\\
- \( x_{ij} \): A binary variable that is 1 if the tour includes a direct path from city \( i \) to city \( j \), and 0 otherwise.\\
$</step>$\\
$<step>$\\
\textbf{Objective Function:}\\
Minimize the total travel cost:
\[
\text{Minimize} \quad \sum_{i} \sum_{j} c_{ij} \cdot x_{ij}
\]\\
$</step>$\\
$<step>$\\
\textbf{Constraints:}\\
1. \textbf{Each city is visited exactly once:}
\[
\sum_{j} x_{ij} = 1 \quad \forall i
\]

2. \textbf{Each city is left exactly once:}
\[
\sum_{i} x_{ij} = 1 \quad \forall j
\]

3. \textbf{Subtour elimination constraints:}
\[
u_i - u_j + n \cdot x_{ij} \leq n - 1 \quad \forall i, j \geq 1, i \neq j
\]
where \( u_i \) is an auxiliary variable representing the position of city \( i \) in the tour, and \( n \) is the number of cities.
\\
$</step>$\\
$<step>$\\
\textbf{Mathematical Model:}
\[
\begin{array}{ll}
\min & \displaystyle \sum_{i} \sum_{j} c_{ij} \cdot x_{ij} \\
\text{s.t.} & \displaystyle \sum_{j} x_{ij} = 1 \quad \forall i \\
& \displaystyle \sum_{i} x_{ij} = 1 \quad \forall j \\
& u_i - u_j + n \cdot x_{ij} \leq n - 1 \quad \forall i, j \geq 1, i \neq j \\
& x_{ij} \in \{0, 1\} \quad \forall i, j
\end{array}
\]\\
$</step>$\\
$<step>$\\
\textbf{Nonlinear Relationships:}\\
No nonlinear relationships are present in this model. This step is unnecessary.
$</step>$\\
$<step>$\\
\textbf{Final Model:}\\
The final linear programming model is defined by the objective function and constraints above.
\\$</step>$
\end{tcolorbox}

\begin{tcolorbox}[title=Python Code Using COPT, colback=orange!5, breakable]
\begin{lstlisting}[language=Python,]
import coptpy as cp
from coptpy import COPT

# Define the cost matrix
cost_matrix = [
[0, 37, 72, 66, 33],
[37, 0, 26, 56, 100],
[72, 26, 0, 49, 32],
[66, 56, 49, 0, 13],
[33, 100, 32, 13, 0]
]

# Number of cities
n = len(cost_matrix)

# Create COPT environment and model
env = cp.Envr()
model = env.createModel("BusTourOptimization")

# Create binary decision variables
x = {}
for i in range(n):
for j in range(n):
    if i != j:
        x[i, j] = model.addVar(vtype=COPT.BINARY, name=f"x_{i}_{j}")

# Create auxiliary variables
u = {}
for i in range(1, n):
u[i] = model.addVar(vtype=COPT.CONTINUOUS, name=f"u_{i}")

# Set objective function
obj = cp.quicksum(
cost_matrix[i][j] * x[i, j] 
for i in range(n) 
for j in range(n) 
if i != j
)
model.setObjective(obj, sense=COPT.MINIMIZE)

# Add constraints: each city visited once
for i in range(n):
model.addConstr(cp.quicksum(x[i, j] for j in range(n) if i != j) == 1)

# Add constraints: each city left once
for j in range(n):
model.addConstr(cp.quicksum(x[i, j] for i in range(n) if i != j) == 1)

# Add subtour elimination constraints
for i in range(1, n):
for j in range(1, n):
    if i != j:
        model.addConstr(u[i] - u[j] + n * x[i, j] <= n - 1)

# Solve the model
model.solve()

\end{lstlisting}
\end{tcolorbox}

\subsection{Qwen3's response}
\begin{tcolorbox}[title=Python Code Using COPT, breakable, colback=orange!5]
\begin{lstlisting}[language=Python]
import coptpy as cp
from coptpy import COPT

env = cp.Envr()
model = env.createModel()

# Define cost matrix
cost = [
[0, 37, 72, 66, 33],
[37, 0, 26, 56, 100],
[72, 26, 0, 49, 32],
[66, 56, 49, 0, 13],
[33, 100, 32, 13, 0]
]

# Create variables
x = [[None for _ in range(5)] for _ in range(5)]
for i in range(5):
for j in range(5):
    if i != j:
        x[i][j] = model.addVar(vtype=COPT.BINARY)

# Objective function
obj_expr = 0
for i in range(5):
for j in range(5):
    if i != j:
        obj_expr += cost[i][j] * x[i][j]
model.setObjective(obj_expr, COPT.MINIMIZE)

# Constraints
for i in range(5):
model.addConstr(sum(x[i][j] for j in range(5) if j != i) == 1)
for i in range(5):
model.addConstr(sum(x[j][i] for j in range(5) if j != i) == 1)

# Solve
model.solve()
\end{lstlisting}
\end{tcolorbox}

\subsection{Discussions on two models' results}
This example clearly shows why rewards should be based on step-by-step reasoning rather than just the optimal value. During the reasoning process, StepORLM correctly recognises the importance of subtour elimination constraints, whereas Qwen3 initially identifies them but then overlooks them due to circular self-debate. In this case, the presence or absence of these constraints does not affect the final result. Therefore, both codes can obtain the correct optimal solution. 

\section{Step-level Reasoning Trajectory Template}
\label{appendix:initial_trajectory}

\begin{tcolorbox}
    [colback=orange!5]

    \begin{enumerate}
        \item \textbf{Problem Description:} Clearly identify the problem, restating
            unclear expressions and defining the optimization objective.

        \item \textbf{Sets and Parameters:} Define all relevant sets (e.g., products,
            time periods, locations) and parameters (e.g., costs, capacities,
            demands).

        \item \textbf{Decision Variables:} Introduce appropriate decision
            variables, specifying their domains and interpretations.

        \item \textbf{Objective Function:} Formulate the mathematical
            expression for the objective to be minimized or maximized.

        \item \textbf{Constraints:} Develop all necessary constraints, explaining
            the reasoning behind each.

        \item \textbf{Mathematical Model Summary:} Present a concise
            mathematical formulation of the complete model.

        \item \textbf{Nonlinear Relationships (if applicable):} Handle any
            nonlinear relationships through transformation or approximation
            techniques.

        \item \textbf{Final Model and Implementation Considerations:}
            Articulate the detailed model with any implementation-specific
            adaptations.
    \end{enumerate}
\end{tcolorbox}

Each reasoning step is enclosed within \texttt{<step>...</step>} tags. While this structure is fixed during the initial warm-up stage, it becomes more flexible in subsequent self-evolving iterations. This is because we do not constrain the model's output format or the number of steps during this phase, allowing the LLM to learn more complex and implicit reasoning patterns

\end{document}